\documentclass[sigconf]{acmart}

\usepackage{xcolor}         % colors
\usepackage{caption}
\usepackage{subcaption}
\usepackage[utf8]{inputenc} % allow utf-8 input
\usepackage[T1]{fontenc}    % use 8-bit T1 fonts
\usepackage{hyperref}       % hyperlinks
\usepackage{url}            % simple URL typesetting
\usepackage{booktabs}       % professional-quality tables
\usepackage{amsmath}
\usepackage{amsfonts}
\usepackage{graphicx}
\usepackage{algorithm}
\usepackage{algcompatible}

\newcommand{\y}{{\bf y}}

\newcommand{\Z}{{\bf Z}}

\DeclareMathOperator*{\argmin}{arg\,min}
\usepackage{pifont}% http://ctan.org/pkg/pifont
\DeclareRobustCommand*\cal{\relax\mathcal}

\copyrightyear{2022}
\acmYear{2022}
\setcopyright{acmcopyright}\acmConference[ICAIF '22]{3rd ACM International Conference on AI in Finance}{November 2--4, 2022}{New York, NY, USA}
\acmBooktitle{3rd ACM International Conference on AI in Finance (ICAIF '22), November 2--4, 2022, New York, NY, USA}
\acmPrice{15.00}
\acmDOI{10.1145/3533271.3561772}
\acmISBN{978-1-4503-9376-8/22/10}

\begin{document}

\title{StyleTime: Style Transfer for Synthetic Time Series Generation}

\author{Yousef El-Laham}
\email{yousef.el-laham@jpmchase.com}
\affiliation{%
  \institution{J.P. Morgan AI Research}
  \streetaddress{}
  \city{New York}
  \state{New York}
  \country{USA}
  \postcode{}
}

% \author{Elizabeth Fons}
% \email{elizabeth.fons@jpmchase.com}
% \affiliation{%
%   \institution{J.P. Morgan AI Research}
%   \streetaddress{}
%   \state{New York}
%   \postcode{}
% }

\author{Svitlana Vyetrenko}
\email{svitlana.vyetrenko@jpmchase.com}
\affiliation{%
  \institution{J.P. Morgan AI Research}
  \streetaddress{}
  \city{New York}
  \state{New York}
  \country{USA}
  \postcode{}
}


\begin{abstract}
 Neural style transfer is a powerful computer vision technique that can incorporate the artistic ``style" of one image to the ``content" of another. The underlying theory behind the approach relies on the assumption that the style of an image is represented by the Gram matrix of its features, which is typically extracted from pre-trained convolutional neural networks (e.g., VGG-19). This idea does not straightforwardly extend to time series stylization since notions of style for two-dimensional images are not analogous to notions of style for one-dimensional time series. In this work, a novel formulation of time series style transfer is proposed for the purpose of synthetic data generation and enhancement. We introduce the concept of stylized features for time series, which is directly related to the time series realism properties, and propose a novel stylization algorithm, called StyleTime, that uses explicit feature extraction techniques to combine the underlying content (trend) of one time series with the style (distributional properties) of another. Further, we discuss evaluation metrics, and compare our work to existing state-of-the-art time series generation and augmentation schemes. To validate the effectiveness of our methods, we use stylized synthetic data as a means for data augmentation to improve the performance of recurrent neural network models on several forecasting tasks.        
\end{abstract}




\maketitle

\section{Introduction}
\label{s: introduction}
Synthetic generation of labeled images has successfully been applied to improve generalization capability, and, therefore, performance of machine learning image detection algorithms. Little work, however, has been done on using synthetic time series to improve generalization capability of machine learning forecasting algorithms. In financial services, for example, there is interest in augmenting historical stock price time series with synthetic rare event data in order to enable price prediction models generalize better on unseen scenarios during the unexpected market shocks due to economic events, global pandemic, flash crashes, etc. . One generally expects models trained on such augmented time series to perform well in practice if synthetic time series is statistically similar to real out-of-sample data as suggested by the ``train on synthetic, test on real" (TSTR) framework \cite{esteban2017realvalued}. Moreover, statistical realism of the augmented time series dataset is often a requirement for explainability purposes in applications such as finance. %due to the regulatory constraints that arise in practice.

Multi-agent simulation provides a means of generating synthetic time series using a bottom-up approach. In multi-agent market simulation, the resultant time series originate from interaction of a large number of market agents (e.g., \cite{abides}). Because of the flexibility to change market configuration on the agent level, multi-agent simulation is well-suited for modeling counterfactual scenarios, however, it is notoriously difficult to calibrate to reflect realistic properties of the market \cite{vyetrenko2019get}.

%\cred{
%\begin{itemize}
%\item Neural style transfer comes from computer vision literature, but has limitations when it come time series
%\item Existing neural style transfer methods for time series obtain style representation using the Gram matrix (add references). We show in this paper that this is not the best approach to extracting style for a time series
%\item In this paper, we address this limitation and demonstrate how style transfer can be used to enhance the realism of synthetic data
%\item We formulate the problem of neural style transfer for time series and motivate its use by a data augmentation task in order to improve the performance of models on various supervised learning tasks
%\end{itemize}
%}
In this paper, we propose to use neural style transfer (NST) to improve statistical realism of synthetic time series, and use it to stylize time series that model counterfactual scenarios for the purpose of data augmentation. NST was initially introduced in \cite{gatys2015neural} to incorporate artistic style into photographs and quickly won over the imaging community. By including the feature correlations of multiple layers, one can obtain a stationary, multi-scale representation of the input image (i.e., the Gram matrix), which captures its texture information but not the global arrangement -- which is called style representation. %The described style representation construction naturally resonates with the intermittency property of stock price time series -- at any micro or macro time scale, asset price returns must display high degree of volatility \cite{vyetrenko2019get}.
%{\color{blue}NST can help disassociate content image from style, and hence generate a dataset with new content-style combinations. By utilizing this idea, NST was used as an augmentation technique to improve quality of machine learning classification algorithms for images \cite{nst_data_aug}.} 
Despite receiving attention by the computer vision community, there is little academic work on NST in the context time series. In \cite{seismic}, NST was used to introduce realistic noise to the seismic shock model. In \cite{financestyle}, denoising autoencoder combined with NST was used as a generative technique for realistic daily price time series. The generated time series showed statistical properties similar to historical data;  additionally, the visual inspection of generated paths indicated presence of technical patterns that are characteristic of the historical daily asset time series. Both of the aforementioned methods extract time series style representations in the same manner they are extracted for images in \cite{gatys2015neural}, which may not be suitable notions of styles for realistic time series since they only contain information about correlations across extracted features. Further related work also includes NST for audio and voice synthesis. This class of NST methods utilize a variety of different techniques for extracting style representations including frequency-domain representations (e.g., VGG-19 or wide-shallow-random networks applied to 2D spectrograms \cite{ustyuzhaninov2016texture}), time-domain representations (e.g., SoundNet encoder \cite{aytar2016soundnet} or WaveNet decoder \cite{oord2016wavenet} applied to raw waveforms), and domain expert knowledge of auditory perception \cite{grinstein2018audio}. Unfortunately, since these style representations are specific to audio data, they cannot be directly applied to other domains, such as finance. 

% \cgreen{WHY SYNTHETIC GENeRATION OF AUDIO DESERVES SPECIAL ATTENTION} \cgreen{For instance, in [Adversarial Audio Synthesis paper] a first attempt at applying GANs to unsupervised synthesis of raw-waveform audio is introduced.}

% \cgreen{Audio style transfer paper}

%Neural Style Transfer for Audio Spectrograms

% \cgreen{In [Fourier flows paper], a flow generative model for time-series in the Fourier domain is presented. The time-series data are first converted to the Fourier domain. }

% The developed approaches in this paper are designed for time series data and extract style representations related to  time series realism properties.
\paragraph{Main Contributions:} We develop a novel framework for time series stylization with the goal of constructing realistic and useful synthetic datasets. For that, we propose a style transfer method that combines the underlying trend of one time series with the distributional properties of another. These distributional properties, which we refer to as \emph{stylized features}, are informed by domain knowledge of time series data and can be computed using sample-based approximations, without requiring the training of dedicated feature extraction techniques like convolutional neural networks (CNNs). Trends are extracted by applying time series decomposition on either the original time series dataset or a synthetic dataset generated from another model. To validate our method, we empirically show on three different datasets that our time series style transfer method is able to achieve competitive performance with state-of-the-art methods in terms of fidelity, predictive utility, and authenticity.

% use both explicit and implicit feature extraction techniques based on GANs as well as Gaussian processes (GPs). We apply the developed algorithms to a financial application, where the goal is in generating realistic synthetic daily price time series data. We use price return distributions as market realism metrics, and demonstrate that by stylizing simple content time series with historical price data, we can construct a dataset that achieves a high degree of realism. Finally, we illustrate the use of stylized time series by augmenting training datasets of several supervised learning algorithms that predict stock price moves, and improve their performance as a result.
\paragraph{Related Work:} 
Realistic time series generation has been previously studied in the literature by using the generative adversarial networks (GANs). With the TimeGAN architecture~\cite{Yoon2019TimeseriesGA}, realistic generation of temporal patterns was achieved by jointly optimizing with both supervised and adversarial objectives to learn an embedding space. QuantGAN \cite{Wiese_2020} consists of a generator and discriminator functions represented by temporal convolutional networks, which allows it to synthesize long-range dependencies such as the presence of volatility clusters that are characteristic of financial time series. TimeVAE \cite{desai2021timevae} was recently proposed as a variational autoencoder alternative to GAN-based time-series generation. In \cite{alaa2020generative}, Fourier flows, a flow-based generative model for time-series in the Fourier domain, is presented. Fourier flows are able to achieve competitive performance with state-of-the-art generation techniques at much lower training times. Methods like GANs, VAEs and Fourier flows are typically used for learning the underlying distribution of the training data, and may not provide the distributionally new scenarios needed for data augmentation. 

% Multi-agent simulation provides a means of generating synthetic time series using the bottom-up approach. For example, in multi-agent market simulation, the resultant time series originate from interaction of a large number of market agents (e.g., \cite{abides}). Because of the flexibility to change market configuration on the agent level, multi-agent simulation is well-suited for modeling counterfactual scenarios, however, it is notoriously difficult to calibrate in order to reflect realistic properties of the market~\cite{vyetrenko2019get}.

% NST can help disassociate content image from style, and hence generate a dataset with new content-style combinations - by utilizing this idea, NST was used as an augmentation technique to improve quality of machine learning classification algorithms for images \cite{nst_data_aug}. Despite receiving attention by the computer vision community, there is little academic work on neural style transfer in the context time series. In \cite{seismic}, neural style transfer was used to introduce realistic noise to the seismic shock model. In \cite{financestyle}, denoising autoencoder combined with the neural style transfer was used as a generative technique for realistic daily price time series. The generated time series showed statistical properties similar to historical data;  additionally, the visual inspection of generated paths indicated presence of technical patterns that are characteristic of the historical daily asset time series.
% {\color{green}to Yousef: one sentence about what will make our approach different from the above papers}
Data augmentation is well established in computer vision tasks due to the simplicity of label-preserving geometric image transformation techniques, but it is still not widely used for time series with some early work being discussed in the literature~\cite{timeseries_augmentation}. For example, simple augmentation techniques applied to financial price time series such as adding noise or time warping were shown to improve the quality of next day price prediction model \cite{fons2021adaptive}, however, such transformations were not required to produce realistic synthetic time series. In the computer vision literature, NST was proposed to disassociate content image from style, and hence generate a dataset with new content-style combinations - by utilizing this idea, NST was used as an augmentation technique to improve quality of machine learning classification algorithms for images \cite{nst_data_aug}. %THIS IS USED TWICE NOW - LET"S AGREE ON A GOOD SPOT FOR IT} 
In this paper, we investigate the use of NST for realistic time series generation and subsequent data augmentation. %We formulate the problem of realistic style transfer for time series and motivate its use by a  synthetic time series augmentation task in order to improve the performance of stock price prediction models.

\section{Background}
\label{s: prior_work}
NST is an algorithm that was introduced in \cite{gatys2015neural} as a means to combine the content of a photograph with the style of an artwork. The theory behind the approach relies on the assumption that CNNs have the capability to capture representations of both the ``content" and the ``style" of an image. Mathematically speaking, NST is an optimization problem, where the goal is to minimize a loss function with respect to an input image $\Z\in{\cal Z}$ (e.g., an RGB image). The loss function consists of two different components: the content loss and the style loss. The content loss ${\cal L}_{\rm c}(\Z, \Z_c)$ penalizes values of $\Z$ that differ from a considered content image $\Z_c\in{\cal Z}$, while the style loss  ${\cal L}_{\rm s}(\Z, \Z_s)$ penalizes values of $\Z$ whose styles differ from a considered style image $\Z_s\in{\cal Z}$. The NST objective function is a weighted sum of the content and style losses:
\begin{equation}
    \label{eq: nst_objective}
    {\cal L}(\Z, \Z_c, \Z_s) = \alpha{\cal L}_{\rm c}(\Z, \Z_c) + \beta{\cal L}_{\rm s}(\Z, \Z_s),
\end{equation}
where the hyperparameters $\alpha, \beta \in \mathbb{R}$ are the content and style weights, respectively. Minimization of this loss function leads yields the stylized image, i.e.,
\begin{equation}
    \label{eq: solving_nst_objective}
    \Z^\star = \argmin_{\Z\in{\cal Z}} {\cal L}(\Z, \Z_c, \Z_s).
\end{equation}
Evaluation of the content and style losses require a method for feature extraction. The most common approach to extracting these features is to use a CNN. The feature representation of the image $\Z$ in the $\ell$th (convolutional) layer of the CNN is given by a rectangular matrix ${\bf F}^\ell(\Z)\in\mathbb{R}^{K_{\ell}\times N_{\ell}}$, whose elements are denoted by
\begin{equation}
\label{eq: feature_matrix_elements}
[{\bf F}^\ell (\Z)]_{i, j}= f_{i, j}^{\ell}(\Z),
\end{equation}
where $i$ denotes the index of the channel and $j$ denotes the index of the feature vector. Given an input image $\Z$ and a content image $\Z_{c}$, the content loss (with respect to the $\ell$th layer of the CNN) is simply the distance between these two images in feature space, i.e., 
\begin{align}
    \label{eq: content_loss}
    {\cal L}_{\rm c}(\Z, \Z_{c}) &= \frac{1}{2}\|{\bf F}^\ell(\Z)-{\bf F}^\ell(\Z_c)\|_{2}^2 \\
    &= \frac{1}{2}\sum_{i=1}^{K_
\ell}\sum_{j=1}^{N_\ell} \left(f_{i, j}^{\ell}(\Z) - f_{i, j}^{\ell}(\Z_c)\right)^2.
\end{align}
To define the style loss, there needs to be a notion of ``style" that can be mathematically represented. In the seminal paper on NST \cite{gatys2015neural}, the authors proposed to extract the style of an image by using the Gram matrix of the CNN features. The elements of the Gram matrix (of the $\ell$th layer), denoted by ${\bf G}_{i, j}^\ell(\Z)$, are computed by taking an inner product across the feature maps, i.e., 
\begin{equation*}
    \label{eq: gram_matrix_elements}
    \hspace{0.5cm} [{\bf G}^\ell(\Z)]_{i,j}=\sum_{k=1}^{N_\ell} f_{i, k}^{\ell}(\Z)f_{j, k}^{\ell}(\Z),  \hspace{-2.75cm}\begin{split}
        &i=1,\ldots,K, \\
        &j=1,\ldots,K.
    \end{split}
\end{equation*}
In matrix form, this operation is equivalent to taking the outer product of the feature map matrix, i.e.,      ${\bf G}^\ell(\Z)= {\bf F}^\ell(\Z){\bf F}^\ell(\Z)^\intercal$.
% \begin{equation}
%     \label{eq: gram_matrix_full}
    %  ${\bf G}^\ell(\Z)= {\bf F}^\ell(\Z){\bf F}^\ell(\Z)^\intercal$
% \end{equation}
Using the Gram matrix representation, discrepancy in style can be computed by evaluating a distance of the Gram matrices of the input image $\Z$ and the style image $\Z_{s}$:
\begin{equation}
    \label{eq: style_loss_layer_l}
    {\cal E}_{\ell}(\Z, \Z_{s}) = \frac{1}{4K_\ell^2N_\ell^2}\|{\bf G}^\ell(\Z)- {\bf G}^\ell(\Z_{s})\|_2^2
\end{equation}
The total style loss is a weighted average of the style losses across each of the convolutional layers of the CNN, i.e.,
\begin{equation}
    \label{eq: total_style_loss}
     {\cal L}_{\rm s}(\Z, \Z_{s}) = \sum_{\ell=1}^L w_l{\cal E}_\ell(\Z, \Z_{s}).
\end{equation}
\section{Problem Formulation}
\label{s: problem_formulation}
The goal of this work is to develop a framework for synthetic data generation and enhancement of time series by using style transfer. Under this framework,  a time series $\y=[y_1,\ldots,y_T]^\intercal\in\mathbb{R}^T$ is optimized such that it captures the overall trend of a content time series $\y_{c}=[y_{c, 1},\ldots,y_{c, T}]^\intercal\in\mathbb{R}^T$ and the realistic distributional properties of a style time series $\y_{s}=[y_{s, 1}, \ldots, y_{s, T}]^\intercal\in\mathbb{R}^T$ (see Fig. 1 for a comparison to image-based style transfer).
\begin{figure}[t]
    \centering
    \includegraphics[trim={0cm, 7.5cm, 14cm, 0cm}, clip, width=0.95\linewidth]{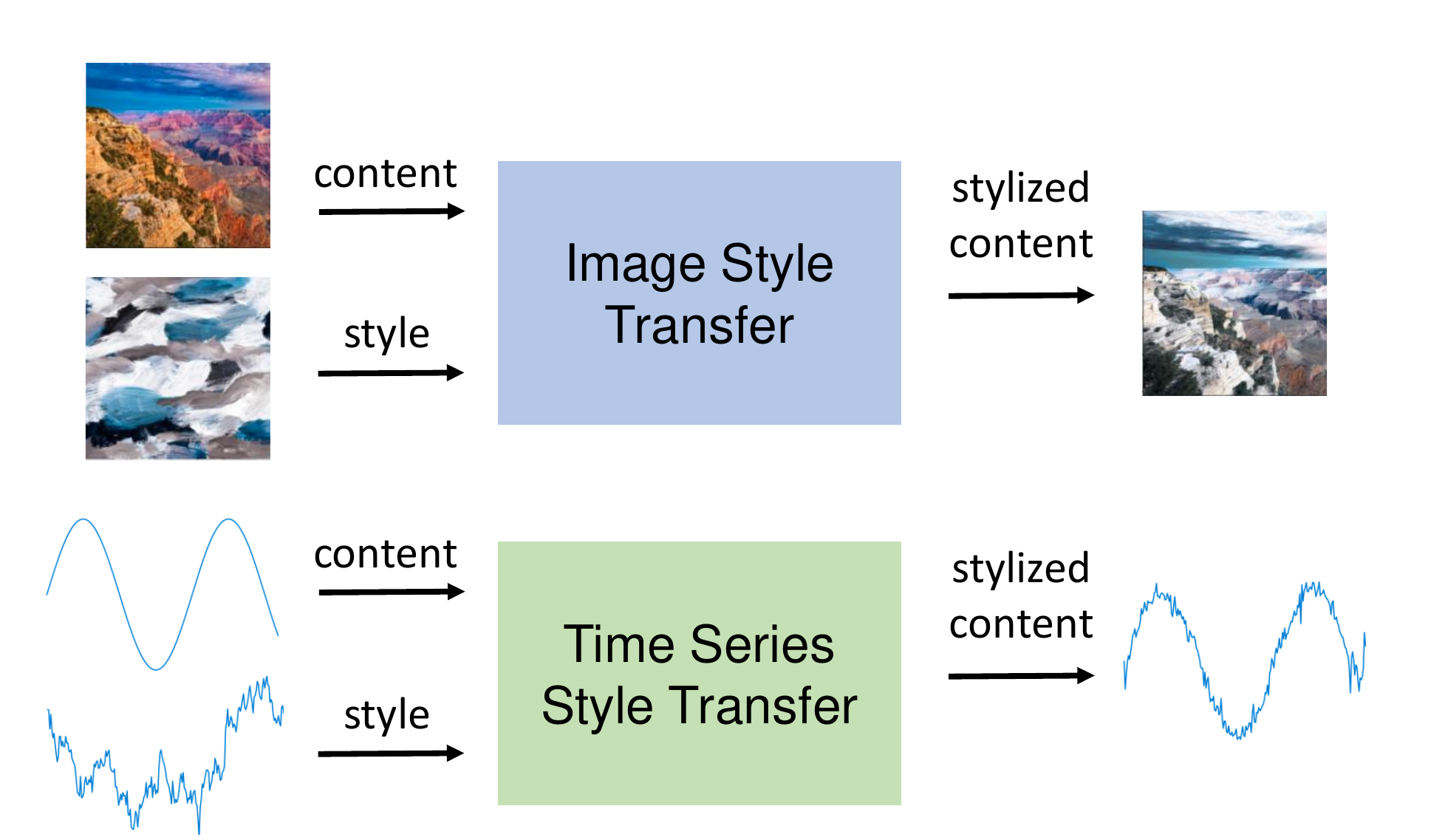}
    \caption{Style transfer for images (top) versus style transfer for time series (bottom).}
    \label{fig: style_transfer_framework}
\end{figure}
One approach to time series stylization is to use the classic NST algorithm outlined in Section \ref{s: prior_work} with a CNN adapted to time series (comprised of one-dimensional convolutional layers) as a means for feature extraction. Such an approach would require that a custom model is trained to solve some task (e.g., time series generation, regression, and/or classification). This approaches is utilized in \cite{financestyle}, where a denoised autoencoder was trained as a means for NST stylization of finanical time series. Another approach uses time series to image conversion techniques and then applies existing image-based NST algorithms. Here, a mapping $g: \mathbb{R}^T\rightarrow{\cal Z}$ can be used to convert a time series into an image. Converted content and style images can then be used to minimize the NST objective function in \eqref{eq: nst_objective} to obtain a solution $\Z^\star$. Finally, the inverse mapping $g^{-1}$ is used to transform the stylized content in the image domain into the time domain to obtain the stylized time series, i.e., $\y^\star=g^{-1}(\Z^\star)$. The benefit of this approach is that pre-trained models can be leveraged for feature extraction (e.g., VGG-19). A variation of this general technique has been successfully applied to stylize synthetic seismic data for realism enhancement \cite{seismic}. An approach similar in spirit that proposes to transform a continuous numeric time-series classification problem to a vision problem has been shown useful for recovering financial signals in \cite{mondrian}.

%\cblue{Alternatively, one can also apply NST directly to visualizations of time series, similar to the approach used in \cite{mondrian} for time series forecasting. }
 
\begin{figure}[t]
    \centering
    \includegraphics[trim={0cm, 5.5cm, 15cm, 0cm}, clip, width=0.95\linewidth]{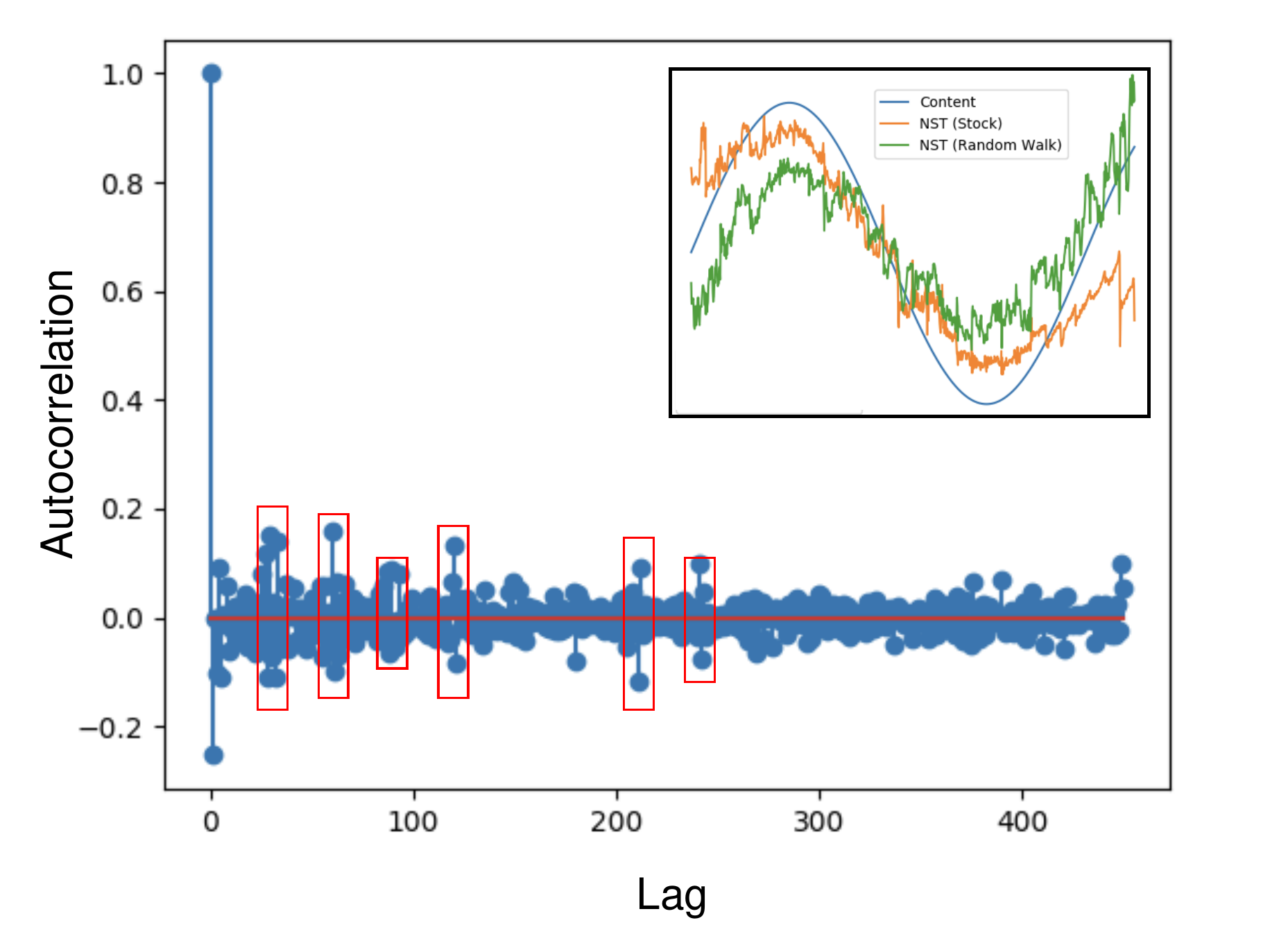}
    \caption{Image-based NST with naive image to time series conversion to stylize a sinusoid with a random walk and a stock price time series . While the resulting stylized time series appear to be reasonable, the autocorrelation function shows spurious spikes at intermittent lags (red boxes).}
    \label{fig: problem_naive_nst}
\end{figure}

Many complications can arise from using the aforementioned approaches. In regards to the time series to image conversion to stylize time series, the mapping $g$ must preserve the correlation structure of the original time series. Naive conversion techniques that construct images by simply arranging the time series spatially can introduce spurious autocorrelations when converting the time series back into the image domain (e.g, see Fig. \ref{fig: problem_naive_nst}).
Another issue is that even if the mapping $g$ is able to preserve the correlation structure of the time series, it must be a bijection (i.e., one-to-one). Conversion techniques such as Markov transition fields and Gramian angular difference fields are non-invertible mappings and thus are automatically excluded from such an approach \cite{wang2015encoding, gamboa2017deep}. Moreover,  time series to image conversion can increase the dimensionality of the data, and hence, lead to increased computational cost of running the style transfer algorithm. For example, while Gramian angular sum fields are invertible mappings that preserve correlation structure of the time series \cite{wang2015imaging}, the range of the mapping is $\mathbb{R}^{T\times T}$. This would prove problematic for time series with a long time horizon $T$. 

Finally, in regards to both of the approaches, there is no evidence to suggest that using the Gram matrix in the image domain extracts a style relevant to realism. To see this, let us denote the column vectors of the feature matrix ${\bf F}^\ell(\y)$ (of the $\ell$th layer of a CNN adapted for time series) by ${\bf f}_1^\ell(\y),\ldots, {\bf f}_{N_{\ell}}^{\ell}(\y)$. If one considers that these column vectors are realizations of some random vector ${\bf f}^\ell(\y)$, then the Gram matrix ${\bf G}^\ell(\y)$ is proportional to an empirical estimator of the second moment of that random vector, i.e.,
\begin{align}
    \label{eq: gram_proportional_second_moment}
    {\bf G}^\ell(\y)&\propto \frac{1}{N_\ell}\sum_{n=1}^{N_{\ell}} {\bf f}_n^\ell(\y){\bf f}_n^\ell(\y)^\intercal \approx \mathbb{E}[{\bf f}^\ell(\y){\bf f}^\ell(\y)^\intercal].
\end{align}
Since the second moment approximation in \eqref{eq: gram_proportional_second_moment} does not capture any other distributional properties, it may not be suitable for capturing the domain-informed stylistic properties of time series.

%First, recall that in the context of computer vision, style typically refers to the artistic style of a photograph or artwork. As we have seen, it has been suggested that information about artistic style is encoded in the Gram matrix of the feature representation of the image. %Unfortunately, there is no evidence to indicate that this representation is useful for time series stylization. 

\section{Style Transfer for Time Series}
\label{s: methodology}
In this paper, we develop a novel style transfer technique, called StyleTime, for synthetic data generation and enhancement of time series. We introduce the concept of \emph{stylized features} as means for style representation, which are features derived from distributional properties of time series that can effectively be used to assess realism. To that end, we formulate StyleTime as an optimization problem, where the content loss is a measure of reconstruction error with respect to underlying trend (content) and the style loss is measure of discrepancy with respect to the stylized features (style).  %To that end, we propose both explicit and implicit stylized feature extraction techniques based on GANs and GP regression. 

Unlike typical generative models, which are capable of generating time series belonging to the domain of training data from which they were trained on, StyleTime is capable of producing realistic time series data with unique trends.  As an example, TimeGAN models trained on a dataset consisting of Brownian motion trajectories is only capable of generating Brownian motion trajectories. In contrast, StyleTime can introduce novel trends outside the domain of the training data. For instance, one would be capable of stylizing a step function with a Brownian motion trajectory. Approaches like this could find applications in rare-event modeling, where one would have the capability of enhancing a time series model representing a hypothetical rare-event (e.g., a market shock due to economic events \cite{mollick2013us}) with the stylized properties of historical data. 

\subsection{Objective Function}
\label{ss: styletime_objective}
StyleTime follows suit to classic style transfer techniques, where features related to content and style are both extracted from two different time series and then an optimization problem is solved to obtain the stylized content. We define the objective function of StyleTime as follows:
\begin{equation}
    \label{eq: StyleTime_obj}
    {\cal L}(\y, \y_c, \y_s) = \alpha {\cal L}_{\rm c}(\y, \y_c) + \beta{\cal L}_{\rm s}(\y, \y_s) + \gamma{\cal L}_{\rm TV}(\y),
\end{equation}
where ${\cal L}_{\rm c}(\y, \y_c)$ denotes the content loss, ${\cal L}_{\rm s}(\y, \y_s)$ denotes the style loss, ${\cal L}_{\rm TV}(\y)$ denotes the total variation loss, and the coefficients $\alpha, \beta, \gamma\in\mathbb{R}^+$ are hyperparameters that determine the relative weight of the content, style, and total variation losses, respectively. We describe each of these losses in the following:

\subsubsection{Content Loss}
\label{sss: style_time_content_loss}
The content loss ${\cal L}_{\rm c}(\y, \y_c)$ measures the discrepancy between the trend of the input time series $\y$ and the trend of the content time series $\y_c$. This can be quantified using the squared $L2$ norm:
\begin{equation}
    \label{eq: StyleTime_content}
    {\cal L}_{\rm c}(\y, \y_c) = \|\y-h(\y_c)\|_2^2,
\end{equation}
where $h(\cdot)$ is a function used to extract the trend of the content time series (e.g., using a moving average filter). In our experiments, we employ an additive seasonal-trend decomposition, where a moving average filter is used in order to extract the trend of the content time series so that the loss in \eqref{eq: StyleTime_content} can be evaluated.

\subsubsection{Style Loss}
\label{sss: style_time_style_loss}
Let ${\bf S}^{\ell}(\y)\in\mathbb{R}^{d_\ell}$ denote the $\ell$th stylized feature of the input $\y$. If we consider $L$ stylized features, we define the style loss as 
\begin{equation}
    \label{eq: StyleTime_style}
    {\cal L}_{\rm s}(\y, \y_s)=\sum_{\ell=1}^L w_\ell \|{\bf S}^{\ell}(\y)-{\bf S}^{\ell}(\y_s)\|_2^2,
\end{equation}
where $w_\ell={1}/{d_\ell}$ for all $\ell=1,\ldots, L$. Here, each stylized feature $S^\ell(\cdot)$ represents a distributional property of the time series data that is related to realism. In this work, we mainly considered three different stylized features. Let ${\bf r}=[r_{1},\ldots,r_{T-1}]^\intercal\in\mathbb{R}^{T-1}$ denote the log returns (i.e., log differences) of the time series $\y$, where each element $r_t$ is defined as follows:
\begin{equation}
    \label{eq: log_returns}
    r_t = \log(y_{t+1}) - \log(y_t), \quad t=1,\ldots,T-1.
\end{equation}
The stylized features are derived from the following quantities:
\begin{enumerate}
    \item {\bf Autocorrelation of Returns}: For a real-valued discrete-time stochastic process $\{X_t: t=1,2,\ldots\}$, the autocorrelation function at lag $\tau$ is given by
    \begin{equation}
        \label{eq: autocorrelation}
        R_X(\tau) = \mathbb{E}[X_{t+\tau}X_t].
    \end{equation}
    The autocorrelation function can be used to measure the correlation between a time-series and itself as a function of the lag $\tau$. In the context of finance, the autocorrelation function of price returns are an important indicator of realism. Therefore, we opt to use the sample autocorrelation function of the log returns as a stylized feature, which can be computed as follows:
    \begin{equation}
        \label{eq: sample_acf}
        \hat{\rho}_\tau({\bf r}) = \frac{\sum_{t=\tau+1}^{T-1}(r_t-\bar{r})(r_{t-\tau}-\bar{r})}{\sum_{t=1}^{T-1} (r_t-\bar{r})^2},
    \end{equation}
    where $\bar{r}=\frac{1}{T-1}\sum_{t=1}^{T-1} r_t$ is the average log return. \\
    \item {\bf Volatility}: In the context of financial time series, volatility is a measure of the standard deviation of the log returns of the price time series. Generally speaking, this quantity can be used to quantify the variation of any given time series over time. Incorporating volatility as a stylized feature facilitates that the underlying noise distribution of the time series being optimized, i.e. $\y$, will have similar variation to that of the style time series $\y_s$. In this work, we use the following estimator of volatility to extract the stylized feature:
    \begin{equation}
        \label{eq: volatility}
        \hat\sigma({\bf r}) = \sqrt{\frac{1}{T-2}\sum_{t=1}^{T-1} (r_t-\bar r)^2}
    \end{equation}
    \item {\bf Power Spectral Density}: In signal processing, the power spectral density (PSD) of a signal is used to measure a signal's strength as a function of frequency. For a discrete-time process, the PSD is simply the discrete Fourier transform (DFT) of the autocorrelation function of the process. In practice, for a time-series $\y$, we can obtain an approximation to the PSD, denoted by $\hat S_f(\y)$, by taking the DFT of the sample autocorrelation:
    \begin{equation}
        \label{eq: psd}
        \hat{S}_f(\y) = \sum_{\tau} \hat{\rho}_\tau(\y) e^{-i(2\pi f)\tau},
    \end{equation}
    where $f$ denotes the frequency component. We use the fast Fourier transform (FFT) algorithm to obtain a fast computation of the DFT. Ultimately, we use the average value (across frequencies) of the sample PSD in \eqref{eq: psd} as the stylized feature.% to be used in StyleTime. 
\end{enumerate}

\subsubsection{Total Variation Loss}
\label{sss: style_time_tv_loss}
In the context of images and videos, the total variation loss was introduced to regularize the optimized image in the NST algorithm for realism enhancement \cite{huang2017real}. By adding this loss to the original NST objective in \eqref{eq: nst_objective}, the resulting stylized content has smoother and more realistic appearance. For time series data, the total variation loss ${\cal L}_{\rm TV}(\y)$ can be used as a means to penalized highly non-stationary behavior (e.g., sudden discontinuities) and is defined as follows:
\begin{equation}
    \label{eq: StyleTime_tv}
    {\cal L}_{\rm TV}(\y) = \sum_{t=1}^{T-1}(y_{t+1}-y_t)^2.
\end{equation}
Empirically, we found that incorporating this regularizer is useful for obtaining more realistic time series, especially in cases where the content time series differs greatly from the style time series. 

\subsection{Implementation}
\label{ss: implementation}
In this subsection, we describe the practical implementation of StyleTime for generating synthetic time series datasets. First, we discuss the implementation of StyleTime for generating a single time series and follow that up with discussion on how one can generate a dataset with mutliple time series.

\subsubsection{StyleTime Block}
\label{sss: styletimeblock}
A summary of the implementation of StyleTime for generating a single time series is shown in Algorithm \ref{alg: StyleTime}. In this implementation, we begin by initializing the value of the optimized variable $\y$ (i.e., the stylized content) as the content time series. Then, at each iteration of the algorithm, we iteratively update the variable $\y$ by using a gradient descent update with respect to the loss function ${\cal L}(\y, \y_c, \y_s)$. In our implementation, we used RMSprop optimzer to complete each gradient step for a fixed number of iterations $I$. Alternatively, one can also run the optimizer until some stopping condition is achieved. 

\begin{algorithm}[t]
\caption{${\tt StyleTime}(\y_c, \y_s)$}
\label{alg: StyleTime}
  \begin{algorithmic}[1]
    \STATEx {\bf Input}: Content time series $\y_c$ and style time series $\y_s$.
    \STATEx {\bf Default Parameters}: $\alpha=1$, $\beta=10$, $\gamma=0.0001$, $I=250$.
    \STATE {\bf Initialization}: Set $\y_1\coloneqq \y_c$.
    \FOR{i=1,\ldots, (I-1)}
    \STATE Evaluate content loss ${\cal L}_c(\y_i, \y_c)$ according to eq. \eqref{eq: StyleTime_content}.
    \STATE Evaluate style loss ${\cal L}_s(\y_i, \y_s)$  according to eq. \eqref{eq: StyleTime_style}.
    \STATE Evaluate total variation loss ${\cal L}_v(\y_i)$  according to eq. \eqref{eq: StyleTime_tv}.
    \STATE Compute total loss as:
    \begin{equation*}
        {\cal L}(\y_i, \y_c, \y_s) = \alpha {\cal L}_{\rm c}(\y_i, \y_c) + \beta{\cal L}_{\rm s}(\y_i, \y_s) + \gamma{\cal L}_{\rm TV}(\y_i)
    \end{equation*}
    \STATE Gradient update of stylized time series:
    \begin{equation*}
        \y_{i+1} \coloneqq \y_{i} - \eta_i \nabla_{\y_i} {\cal L}(\y_i, \y_c, \y_s)
    \end{equation*}
    \STATE Determine the next learning rate $\eta_{i+1}$.
    \ENDFOR
    \STATEx {\bf Return}: $\y_{I}$
  \end{algorithmic}
\end{algorithm}

\subsubsection{Generating Synthetic Time Series Datasets}
\label{sss: styletime_datasets}
We can use Algorithm \ref{alg: StyleTime} to generate a new time series $\tilde\y$ given two other time series $\y_c$ and $\y_s$. Given a historical ``style" dataset ${\cal D}_s=\{\y_s^{(n)}\}_{n=1}^{N_s}$ and a reference ``content" dataset  ${\cal D}_c=\{\y_c^{(n)}\}_{n=1}^{N_c}$, we can generate a synthetic dataset $\widetilde{\cal D}=\{\tilde \y^{(n)}\}_{n=1}^N$ by iteratively applying StyleTime to random samples obtain from ${\cal D}_s$ and ${\cal D}_c$ (see Algorithm \ref{alg: StyleTimeDataset}).

\begin{algorithm}[t]
\caption{Generation of Stylized Time Series Datasets}
\label{alg: StyleTimeDataset}
  \begin{algorithmic}[1]
    \STATEx {\bf Input}: Content dataset ${\cal D}_c$ and style dataset ${\cal D}_s$.
    \STATE {\bf Initialization}: Set $\widetilde{\cal D}\coloneqq \emptyset$
    \FOR{n=1,\ldots, N}
    \STATE Randomly sample content time series $\y_c\sim{\cal D}_c$.
    \STATE Randomly sample style time series $\y_s\sim{\cal D}_s$.
    \STATE Run StyleTime on the sampled time series
    \begin{equation*}
        \tilde\y^{(n)} \gets {\tt StyleTime}(\y_c, \y_s)
    \end{equation*}
    \STATE Append $\tilde\y^{(n)}$ to $\widetilde{\cal D}$.
    \ENDFOR
    \STATEx {\bf Return}: $\widetilde{\cal D}$
  \end{algorithmic}
\end{algorithm}

In practice, we choose our style dataset ${\cal D}_s$ to be the training dataset, while the content dataset can be chosen in multiple ways. In this work, we consider the following approaches:
\begin{enumerate}
    \item {\bf Training dataset}: The content dataset can be also be chosen to be the training dataset. In this case, the synthetic dataset generated via Algorithm \ref{alg: StyleTimeDataset} should be statistically similar to that of the original training dataset.
    \item {\bf Perturbed training dataset}: The content dataset can be chosen to be a ``shocked" version of the training dataset. For example, one can add a randomly shifted and scaled unit-step function to each example in the training dataset. 
    \item {\bf Generative model}: In principle, one can also use StyleTime to enhance the quality of synthetic data generated from generative model. This can be done by using Algorithm \ref{alg: StyleTimeDataset} with content dataset obtained by generating synthetic data from another generative model (e.g., Fourier flows). 
\end{enumerate}

\begin{table*}[t]
\resizebox{\linewidth}{!}{\centering
\begin{tabular}{c|cc|cc|cc|}
\cline{2-7}
                                    & \multicolumn{2}{c|}{\textbf{Switching AR(1)}}                   & \multicolumn{2}{c|}{\textbf{Stock}}                  & \multicolumn{2}{c|}{\textbf{Energy}}                 \\ \cline{2-7} 
                                    & \multicolumn{1}{c|}{$F$-score}      & MAE            & \multicolumn{1}{c|}{$F$-score}      & MAE            & \multicolumn{1}{c|}{$F$-score}      & MAE            \\ \hline
% \multicolumn{1}{|c|}{TimeGAN}       & \multicolumn{1}{c|}{(0.951, 0.011)} & (0.135, 0.058) & \multicolumn{1}{c|}{(0.967, 0.011)} & (0.027, 0.011) & \multicolumn{1}{c|}{(0.748, 0.403)} & (0.052, 0.007) \\ \hline
\multicolumn{1}{|c|}{Fourier flows} & \multicolumn{1}{c|}{$0.9464 \pm 0.0023$} & $0.0073 \pm 0.0007$ & \multicolumn{1}{c|}{$0.9813 \pm 0.0010$} & $0.0079\pm0.0022$ & \multicolumn{1}{c|}{$0.9170\pm0.0075$} & $0.0312\pm0.0004$ \\ \hline
\multicolumn{1}{|c|}{StyleTime (In-Sample)} & \multicolumn{1}{c|}{${\bf 0.9970}\pm{\bf 0.0003}$} & $0.0071\pm0.0009$ & \multicolumn{1}{c|}{$0.9967\pm 0.0001$} & $0.0059\pm0.0009$ & \multicolumn{1}{c|}{$0.9969\pm 0.0003$} & ${\bf 0.0286}\pm {\bf 0.0004}$ \\ \hline
\multicolumn{1}{|c|}{StyleTime (Perturbed)} & \multicolumn{1}{c|}{$0.9962\pm0.0003$} & ${\bf 0.0069}\pm{\bf 0.0006}$ & \multicolumn{1}{c|}{${\bf 0.9971}\pm{\bf 0.0002}$} & ${\bf 0.0057}\pm{\bf 0.0010}$ & \multicolumn{1}{c|}{${\bf 0.9974}\pm {\bf 0.0004}$} & $0.0287\pm0.0001$ \\ \hline
\multicolumn{1}{|c|}{StyleTime + Fourier flows} & \multicolumn{1}{c|}{$0.9948\pm0.0009$} & $0.0077\pm0.0015$  & \multicolumn{1}{c|}{$0.9827\pm0.0099$} & $0.0089\pm0.0022$ & \multicolumn{1}{c|}{$0.9755\pm0.0020$} & $0.0353\pm0.0013$  \\ \hline\hline
\multicolumn{1}{|c|}{{\bf Percent Improvement}} & \multicolumn{1}{c|}{5.35\%} & 5.55\%  & \multicolumn{1}{c|}{1.61\%} & 27.85\% & \multicolumn{1}{c|}{8.77\%} & 8.33\%  \\ \hline
\end{tabular}}
\caption{Assessment of synthetic data fidelity for considered sine, stock, and energy datasets. Metrics are reported as (mean $\pm$ standard error) averaged over 5 different random seeds. Percentage improvements are computed to compare the best performing StyleTime implementation and the Fourier flows baseline for each metric separately.}
\label{tab: exp_p2}
\end{table*}

\section{Experimental Results}
\label{s: experiments}
Here, we show some experiments to demonstrate the validity of StyleTime for time series synthetic data generation. Our evaluation is conducted across three different datasets: 
\begin{enumerate}
    \item \emph{Switching AR(1)}: a synthetic switching first-order autoregression dataset generated via the following model:
    \begin{equation*}
        y_t =  \begin{cases} 
      \alpha_{1, 1} y_{t-1} + \alpha_{1, 0} + \epsilon_t, & t=1, \ldots, T_s-1, \\
      \alpha_{2, 1} y_{t-1} + \alpha_{2, 0} + \epsilon_t, & t=T_s, \ldots, T,
    \end{cases}
    \end{equation*}
    where $y_0\sim{\cal N}(0, 1)$, 
    $T_s=\lfloor 0.8\times T \rfloor$, and $\epsilon_t\overset{i.i.d.}{\sim} {\cal N}(0, 1)$ for all $t$. The parameters of the model are set as follows: $\alpha_{1,0}=0.01$, $\alpha_{1,1}=1.001$, $\alpha_{2,0}=-0.01$, and  $\alpha_{2,1}=0.999$.  This synthetic dataset is intended to model a regime switch that is characteristic of financial data.
    \item \emph{Stock}: univariate time series extracted from the Google stock price historical data. 
    \item \emph{Energy}: univariate time series taken from the UCI energy dataset that contains observations of energy use of appliances. This dataset and the Stock dataset were both used to validate other time series deep generative models, including TimeGAN and Fourier flows \cite{alaa2020generative, Yoon2019TimeseriesGA}.
\end{enumerate}
To convert the above univariate time series to a dataset, we apply the sliding window technique using a window size $W$, which gives us a dataset $\{\y^{(n)}\}_{n=1}^{T-W}$ with $\y^{(n)}=[y_n, \ldots, y_{n+W}]^\intercal\in\mathbb{R}^W$ for $n=1,\ldots,T-W$. For all datasets, we use a sliding window size of $W=30$ and a time horizon of $T=3030$, which leads to each dataset containing $N=3000$ samples. Each dataset is then split into a training set and a test set, where the training set contains the first $N_{train}=2400$ samples and the test set contains the last $N_{test}=600$ samples.

We evaluate the performance of all synthetic data generation techniques from two different angles:
\begin{enumerate}
    \item \emph{Fidelity}: We evaluate the fidelity of style transfer enhanced synthetic data in comparison to one of the state-of-the-art deep generative time-series models (i.e., Fourier flows). Synthetic data fidelity is quantified using precision/recall metrics from \cite{sajjadi2018assessing} and by using the TSTR framework \cite{esteban2017real} to assess the predictive utility of the synthetic time series. 
    \item \emph{Effectiveness}:  We evaluate the predictive utility of the enhanced synthetic data in the context of a data augmentation. Here, the synthetic data generated by each technique is used to augment the original training dataset. The goal is to assess the usefulness of the synthetic data to predict out-of-sample sequences (i.e., to improve predictive performance on the test set). In addition to comparing to synthetic data generated by Fourier flows, we also compare to time series-based data augmentation techniques \cite{wen2020time, fons2020evaluating}.
\end{enumerate}

For StyleTime, we generate synthetic data using Algorithm \ref{alg: StyleTimeDataset} with the default parameter values in Algorithm \ref{alg: StyleTime}. In each of the constructed synthetic datasets, the training dataset was used as the style dataset. We consider three different content datasets: (1) training data, (2) perturbed training data, and (3) Fourier flows synthetic data.
% \subsection{Comparison of Style Transfer Techniques}
% \label{ss: experiments_p1}

% \paragraph{Metrics:} \cblue{XXX.}

% \paragraph{Baselines:} \cblue{XXX.}

\subsection{Assessment of Synthetic Data Fidelity}
\label{ss: experiments_p2}
We assess the fidelity of the synthetic data using two summary metrics: the $F$-score and the mean absolute error (MAE) on a TSTR example. First, we use the framework established in \cite{sajjadi2018assessing} to compute a $\alpha$-precision, $\beta$-recall score, as is done in \cite{alaa2020generative}. The $\alpha$-precision metric measures how much the support of the real data covers the support of the synthetic data (i.e., realism), while the $\beta$-recall metric measures how much the support of the synthetic data covers the support of the real data (i.e., diversity). The two scores are then summarized into a single metric (i.e., the $F$-score) by taking the harmonic mean.  Secondly, we evaluate the quality of synthetic data by assessing its predictive utility on the training data. In particular, for each method, we use synthetic data to train a vanilla recurrent neural network (RNN) model with two LSTM layers, each with 100 hidden units. We then compute the mean absolute error (MAE) of the model on the original training dataset. If the RNN trained on synthetic data generalizes well to the training data, then it is considered to be a high fidelity dataset. We compare StyleTime with Fourier flows, a deep flow-based model shown to be capable of achieving state-of-the-art performance with respect to both the F-score and TSTR MAE. We also compute the metrics for the content time-series extracted for each of the considered datasets.

\begin{figure*}[t]
     \centering
     \begin{subfigure}[b]{0.33\textwidth}
         \centering
         \includegraphics[width=\textwidth]{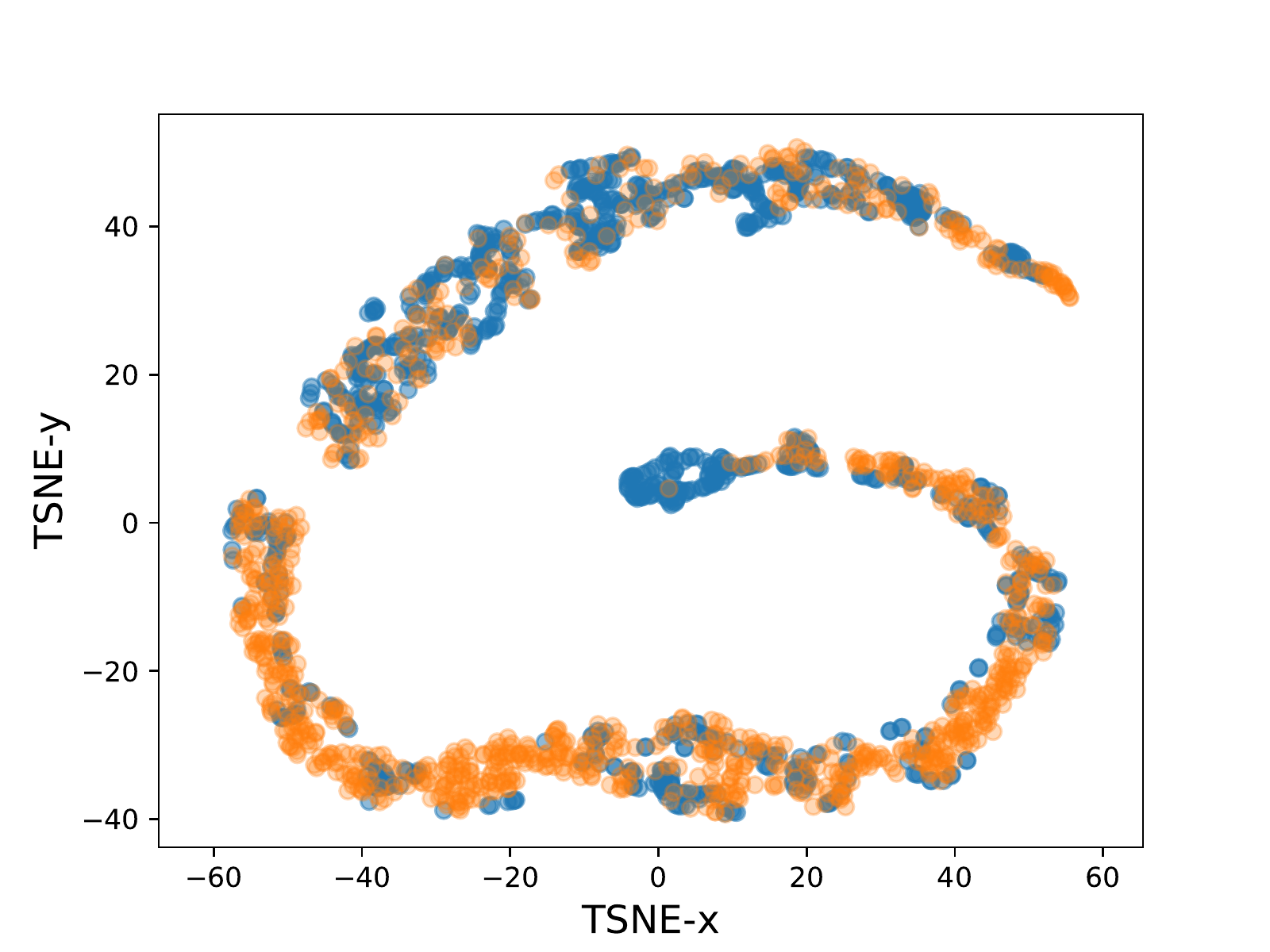}
         \caption{Fourier flows.}
     \end{subfigure}
     \hfill
     \begin{subfigure}[b]{0.33\textwidth}
         \centering
         \includegraphics[width=\textwidth]{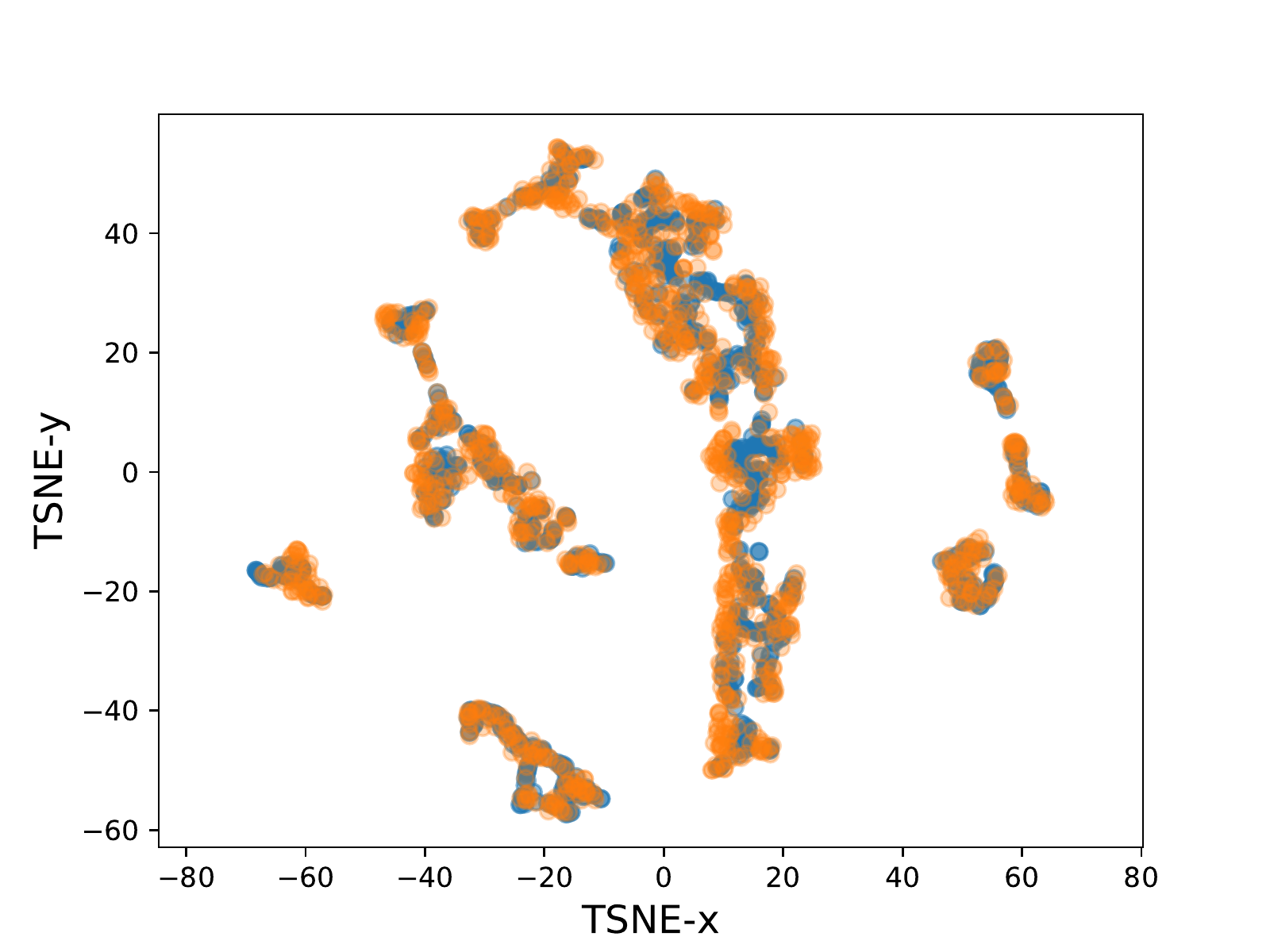}
         \caption{StyleTime (In-Sample).}
     \end{subfigure}
     \hfill
     \begin{subfigure}[b]{0.33\textwidth}
         \centering
         \includegraphics[width=\textwidth]{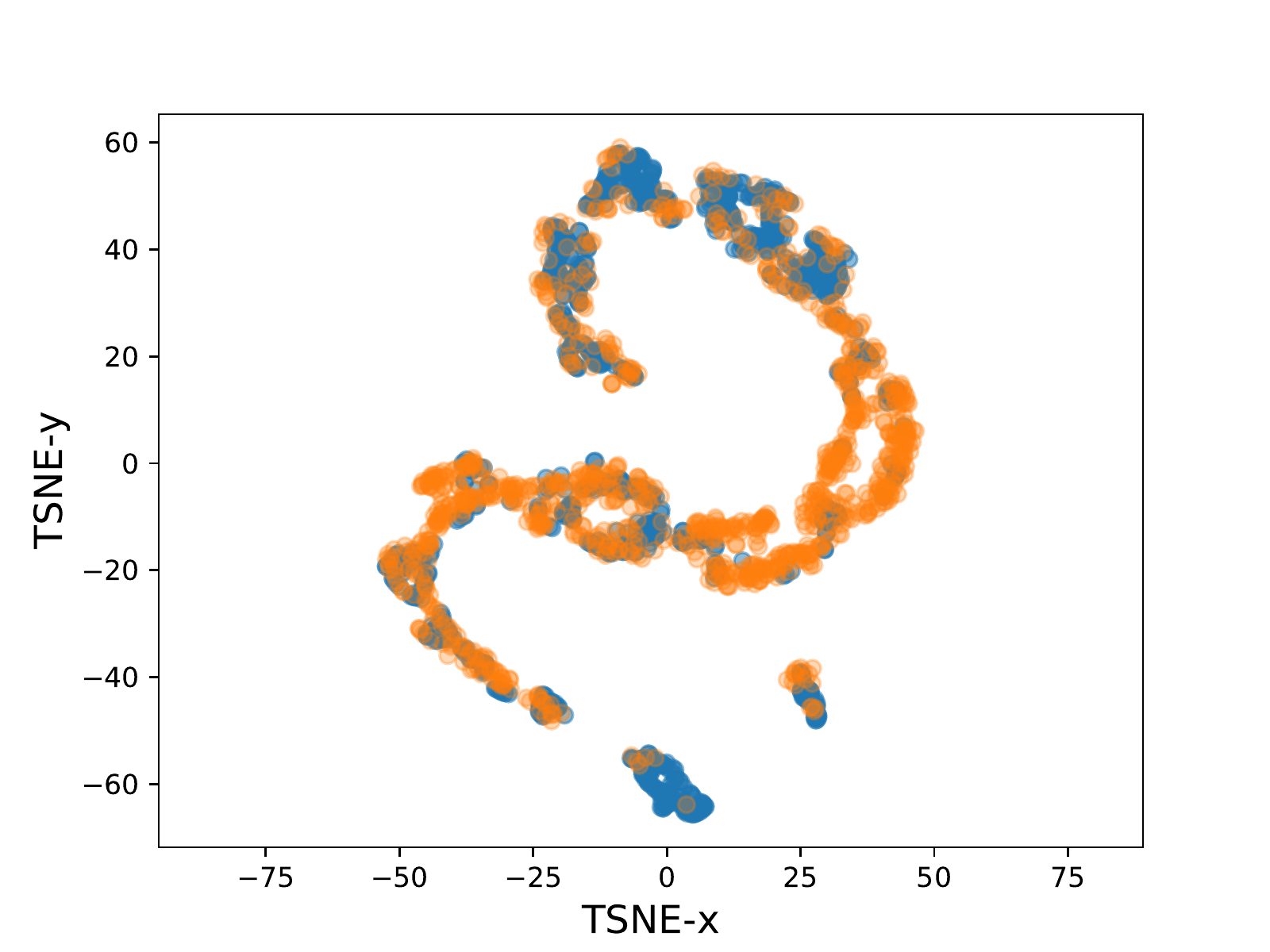}
         \caption{StyleTime + Fourier flows.}
     \end{subfigure}
     \caption{TSNE plots comparing a low dimensional embedding of the Stock training data (blue) with that of the synthetic data generation techniques (orange). }
     \label{fig: tsne_plots_stock}
\end{figure*}

\begin{figure*}[t]
     \centering
     \begin{subfigure}[b]{0.33\textwidth}
         \centering
         \includegraphics[width=\textwidth]{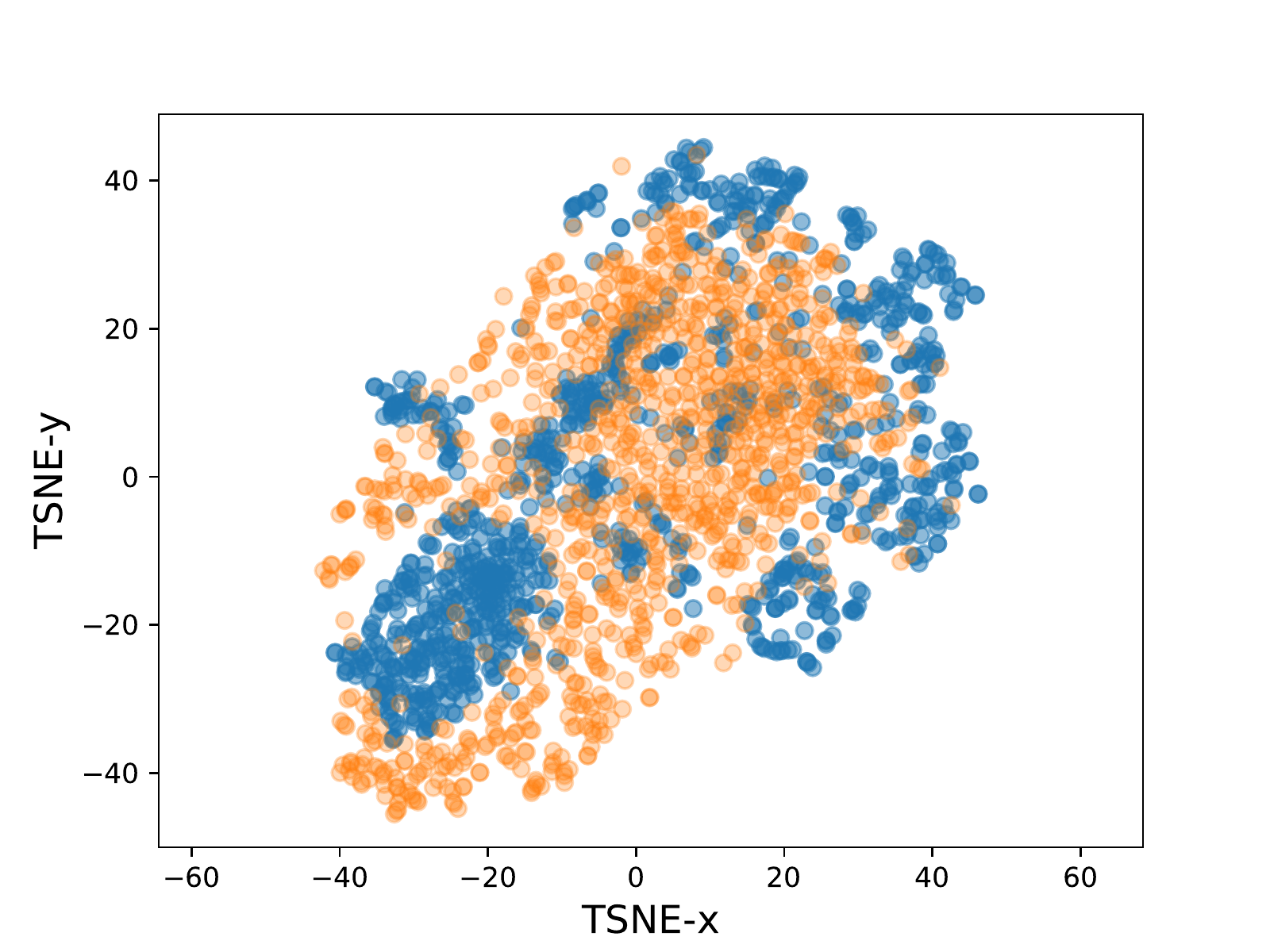}
         \caption{Fourier flows.}
     \end{subfigure}
     \hfill
     \begin{subfigure}[b]{0.33\textwidth}
         \centering
         \includegraphics[width=\textwidth]{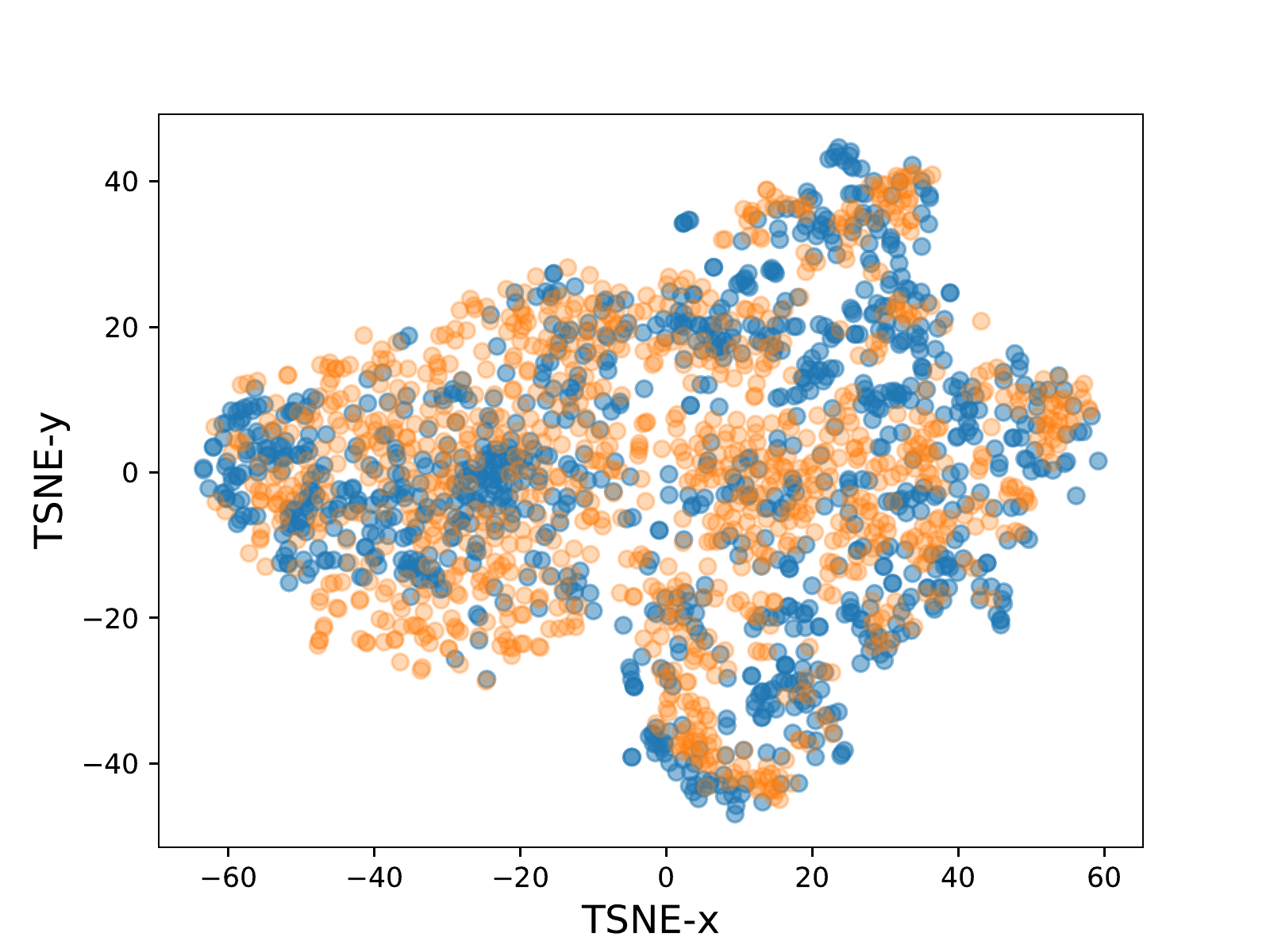}
         \caption{StyleTime (In-Sample).}
     \end{subfigure}
     \hfill
     \begin{subfigure}[b]{0.33\textwidth}
         \centering
         \includegraphics[width=\textwidth]{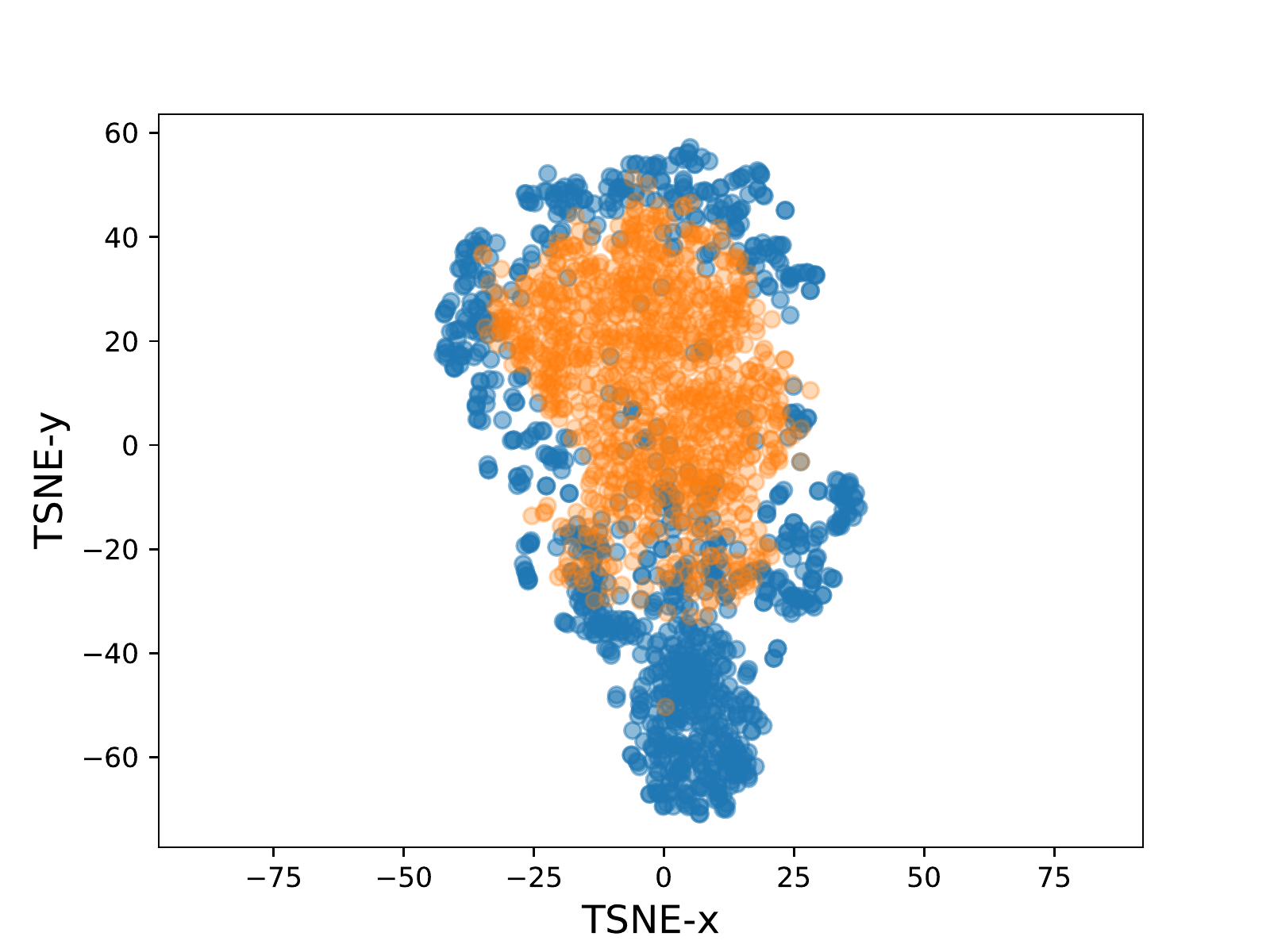}
         \caption{StyleTime + Fourier flows.}
     \end{subfigure}
     \caption{TSNE plots comparing a low dimensional embedding of the Energy training data (blue) with that of the synthetic data generation techniques (orange). Mismatch between training data and Fourier flows synthetic data leeds to low quality stylized synthetic data (StyleTime + Fourier flows). Using  training data directly as content, however, leads to high fidelity samples.} %synthetic data (StyleTime (In-Sample)).}
     \label{fig: tsne_plots_energy}
\end{figure*}

We summarize the results of this experiment in Table \ref{tab: exp_p2}. Across of each of the datasets, synthetic data generated via StyleTime outperforms Fourier flows considering both metrics. Unsurprisingly, StyleTime variations that use in-sample and perturbed content give the best results in terms of both F-score and MAE, while the variation that uses Fourier flows content actually degrades performance with respect to the Fourier flows baseline. This could be an artifact in the quality of the Fourier flows synthetic data used as the content dataset, rather than StyleTime technique itself. This is evidenced by the TSNE plots in Fig. \ref{fig: tsne_plots_energy}, where we can see that the Fourier flow synthetic data does not adequately learn the distribution of the energy data, possibly due to the datasets non-stationary nature. A similar result is also implied for the Stock dataset when looking at the TSNE plots in Fig. \ref{fig: tsne_plots_stock}, where we again see that using in-sample content ultimately leads to higher fidelity synthetic data than other techniques.

% \begin{figure}[t]
%     \centering
%     \includegraphics[trim={0cm, 0cm, 0cm, 0cm}, clip, width=\linewidth]{figures/tsne_energy_30_ff.pdf}
%     \caption{.}
%     \label{fig: tsne_ff_energy}
% \end{figure}
% \begin{figure}[t]
%     \centering
%     \includegraphics[trim={0cm, 0cm, 0cm, 0cm}, clip, width=\linewidth]{figures/tsne_energy_30_nst_standard.pdf}
%     \caption{.}
%     \label{fig: tsne_nst_energy}
% \end{figure}
% \begin{figure}[t]
%     \centering
%     \includegraphics[trim={0cm, 0cm, 0cm, 0cm}, clip, width=\linewidth]{figures/tsne_energy_30_nst_ff.pdf}
%     \caption{.}
%     \label{fig: tsne_nst_ff_energy}
% \end{figure}

\subsection{Assessment of Predictive Utility}
\label{ss: experiments_p3}
In the previous experiment, we tested the fidelity of the synthetic data using a TSTR framework. Unfortunately, TSTR does not provide a means to assessing how well the generation methods extrapolate the underlying distribution of the real dataset. Moreover, the MAE in the previous subsection was computed in absence of a baseline (i.e., train on real, test on real). Here, we test the predictive utility of the synthetic data in the context of data augmentation. For each model, we augment the training data for a variety of different levels (i.e., augmentation size) and assess the MAE of the previous vanilla RNN model for predicting the test dataset. For each method, we report the MAE at the best augmentation level. To test the novelty of the synthetic data, we also apply the authenticity metric proposed in \cite{alaa2021faithful} to each proposed data augmentation method.\footnote{Our implementation of the authenticity metric utilizes the euclidean distance to measure the discrepancy between real and synthetic samples, which is different than what was proposed in \cite{alaa2021faithful}.} The authenticity metric (denoted by $\hat{A}\in[0, 1]$) measures the degree to which the synthetic dataset is novel with respect to the training data. A high authenticity score ($\hat{A}\approx1$) implies that the learned generative model that produced the synthetic data produces authentic (and unseen) time series that do not appear in the training data. A low authenticity score $(\hat{A}\approx0)$ implies that the learned generative model is just producing a noisy copy of data appearing in the training dataset. Intuitively, to achieve better generalization and improve predictive utility, successful time series augmentation is expected to have out-of-distribution properties and, hence, display a high authenticity score. Ideally, synthetic data should have a high authenticity score, as producing copies of the data could give rise to privacy concerns. Moreover, a low authenticity score can also be a good technique for assessing whether or not the generation technique has overfit to the training data. For baselines, we compare to Fourier flows and data augmentation techniques designed for time-series (e.g., jittering, flipping, and time warping \cite{fons2020evaluating}). We summarize the results of this experiment in Table \ref{tab: exp_p3}. Results indicate that StyleTime is able to achieve competitive (but not better) performance with respect to the baselines  across both the augmentation MAE metric (AugMAE) and the authenticity metric. Importantly, StyleTime is able to achieve decent augmentation performance, while maintaining a high authenticity score with respect to the other baselines. For example, while the time warping augmentation method gives the best AugMAE, its authenticity score is significantly lower than that of all variations of StyleTime.

\begin{table*}[t]
\resizebox{\linewidth}{!}{\centering
\begin{tabular}{c|cc|cc|cc|}
\cline{2-7}
                                    & \multicolumn{2}{c|}{\textbf{Switching AR(1)}}                   & \multicolumn{2}{c|}{\textbf{Stock}}                  & \multicolumn{2}{c|}{\textbf{Energy}}                 \\ \cline{2-7} 
                                    & \multicolumn{1}{c|}{AugMAE}   & $\hat{A}$         & \multicolumn{1}{c|}{AugMAE}   & $\hat{A}$            & \multicolumn{1}{c|}{AugMAE}   & $\hat{A}$            \\ \hline
% \multicolumn{1}{|c|}{Baseline}       & \multicolumn{1}{c|}{(0.0167, 0.0017)} & - & \multicolumn{1}{c|}{(0.0055, 0.0010)}  & - & \multicolumn{1}{c|}{(0.0270, 0.0006) } & - \\ \hline
% \multicolumn{1}{|c|}{TimeGAN}       & \multicolumn{1}{c|}{(0.0228, 0.0048)} & (0.961, 0.066) & \multicolumn{1}{c|}{(0.0077, 0.0009)}  & (1.000, 0.000) & \multicolumn{1}{c|}{(0.0283, 0.0015)} & (0.798, 0.294) \\ \hline
\multicolumn{1}{|c|}{Fourier flows} & \multicolumn{1}{c|}{$0.0089 \pm 0.0008$} & $0.9954\pm0.0013$ & \multicolumn{1}{c|}{$0.0057\pm0.0011$} & $0.9760\pm0.0025$ & \multicolumn{1}{c|}{$0.0265\pm0.0004$ } & $0.9942\pm0.0015$ \\ \hline
% \multicolumn{1}{|c|}{Bootstrap} & \multicolumn{1}{c|}{(X.XXX, X.XXX)} & --- & \multicolumn{1}{c|}{(0.0051, 0.0006)} & --- & \multicolumn{1}{c|}{(0.0272, 0.0010)} & --- \\ \hline
\multicolumn{1}{|c|}{Flip} & \multicolumn{1}{c|}{$0.0086 \pm 0.0007$} & $0.9958 \pm 0.0000$ & \multicolumn{1}{c|}{$0.0062\pm0.0005$} & $0.9475\pm 0.0000$ & \multicolumn{1}{c|}{$0.0297\pm0.0007$} & ${\bf 1.0000}\pm {\bf 0.0000}$ \\ \hline
\multicolumn{1}{|c|}{Jitter} & \multicolumn{1}{c|}{$0.0086\pm0.0016$} & $0.9991\pm0.0004$ & \multicolumn{1}{c|}{$0.0057\pm0.0010$} & $0.8665\pm0.0029$ & \multicolumn{1}{c|}{$0.0267\pm0.0007$} & $0.2979\pm0.0044$ \\ \hline
\multicolumn{1}{|c|}{Time Warp} & \multicolumn{1}{c|}{${\bf 0.0081}\pm {\bf 0.0008}$} & $0.6349\pm0.0127$ & \multicolumn{1}{c|}{$0.0054\pm0.0006$} & $0.5981\pm0.0056$ & \multicolumn{1}{c|}{$0.0267\pm0.0006$} & $0.5871\pm0.0167$ \\ \hline
\multicolumn{1}{|c|}{StyleTime (In-Sample)} & \multicolumn{1}{c|}{$0.0084\pm0.0010$} & $0.9986\pm0.0010$ & \multicolumn{1}{c|}{$0.0055\pm0.0004$} & $0.9733\pm0.0028$ & \multicolumn{1}{c|}{$0.0264\pm0.0003$} & $0.8761\pm0.0085$ \\ \hline
\multicolumn{1}{|c|}{StyleTime (Perturbed)} & \multicolumn{1}{c|}{$0.0082\pm0.0008$} & ${\bf 0.9997}\pm{\bf 0.0002}$ & \multicolumn{1}{c|}{${\bf 0.0054}\pm{\bf 0.0004}$} & ${\bf 0.9943}\pm{\bf 0.0017}$ & \multicolumn{1}{c|}{${\bf 0.0263}\pm{\bf 0.0005}$} & $0.8787\pm0.0077$ \\ \hline
\multicolumn{1}{|c|}{StyleTime + Fourier flows} & \multicolumn{1}{c|}{$0.0091\pm0.0008$} & $0.9994\pm0.0004$ & \multicolumn{1}{c|}{$0.0058\pm0.0013$} & $0.9879\pm0.0013$ & \multicolumn{1}{c|}{$0.0264\pm0.0004$} & $0.9441\pm0.0035$ \\ \hline\hline
\multicolumn{1}{|c|}{{\bf Percent Improvement}} & \multicolumn{1}{c|}{-1.23\%} & 0.06\%  & \multicolumn{1}{c|}{0.00\%} & 1.87\% & \multicolumn{1}{c|}{1.5\%} & -5.59\%  \\ \hline
\end{tabular}}
\caption{Assessment of synthetic data predictive utility and authenticity for considered sine, stock, and energy datasets. Metrics are reported as (mean, standard error) averaged over 5 different random seeds. Percentage improvements are computed to compare the best performing StyleTime implementation and the best performing baseline for each metric separately. }
\label{tab: exp_p3}
\end{table*}

\section{Conclusions and Future Work}
\label{s: conclusion}
In this paper, we proposed a new formulation of style transfer for realistic time series generation and enhancement. We explained the reason why style transfer for time series deserves attention as an independent problem and justified that existing NST techniques for vision do not directly transfer to time series space. We proposed a novel time series specific stylization algorithm, called StyleTime, that uses explicit feature extraction techniques to combine the content (i.e., trend) of one time series with the style (i.e., distributional properties) of another. We show that StyleTime is able generate synthetic data that achieves a high degree of realism with regard to a variety of metrics. Furthermore, we demonstrated that synthetic time series stylized by StyleTime can augment training datasets to improve the predictive performance of recurrent neural network forecasting models. In the future, we are looking to expand our work to designing methods for stylization of multi-dimensional time series when both time- and cross-series correlations need to be accounted for. We expect such methods to be applicable to realistic enhancement of synthetic limit orderbook datasets, which can in turn be used for data augmentation of market price and volume prediction algorithms on the microstructure level. %Data augmentation for time series is a new rapidly growing field, along with the style transfer for time series 
% We also plan to investigate automated methods of synthetic time series augmentation in order to maximize the impact of stylization.

\begin{acks}
This paper was prepared for informational purposes  by the Artificial Intelligence Research group of JPMorgan Chase \& Co and its affiliates (“J.P. Morgan”), and is not a product of the Research Department of J.P. Morgan.  J.P. Morgan makes no representation and warranty whatsoever and disclaims all liability, for the completeness, accuracy or reliability of the information contained herein.  This document is not intended as investment research or investment advice, or a recommendation, offer or solicitation for the purchase or sale of any security, financial instrument, financial product or service, or to be used in any way for evaluating the merits of participating in any transaction, and shall not constitute a solicitation under any jurisdiction or to any person, if such solicitation under such jurisdiction or to such person would be unlawful.    
\end{acks}

\bibliographystyle{ACM-Reference-Format}
\bibliography{references/references}

% \appendix









\end{document}